\documentclass[10pt,twocolumn,letterpaper]{article}

\usepackage[pagenumbers]{cvpr} 

\definecolor{cvprblue}{rgb}{0.21,0.49,0.74}
\usepackage[pagebackref,breaklinks,colorlinks,allcolors=cvprblue]{hyperref}
\usepackage{float}
\usepackage{caption}
\usepackage{hyperref}
\usepackage{url}
\usepackage{graphicx}
\usepackage{subcaption}
\usepackage{booktabs}
\usepackage{multirow}
\usepackage{wrapfig}
\usepackage{booktabs}
\usepackage{wrapfig}
\usepackage{bbding}
\usepackage{etoc}

\def\paperID{} 
\def\confName{CVPR}
\def\confYear{2026}

\title{Towards Photorealistic and Efficient Bokeh Rendering via Diffusion Framework}

\author{
  Linxiao Shi$^{1,2}$\thanks{Equal contribution. Intern at vivo BlueImage Lab.} ~~ 
  Siming Zheng$^2$\footnotemark[1] ~~ Zerong Wang$^2$ ~~ Hao Zhang$^2$ ~~ Jinwei Chen$^2$ ~~ Bo Li$^2$ ~~ \\
  Shifeng Chen$^{1,3}$\footnotemark[2] ~~ 
  Peng-Tao Jiang$^2$\thanks{Corresponding author.} \\
$^1$Shenzhen Institutes of Advanced Technology, Chinese Academy of Sciences\\
$^2$vivo BlueImage Lab, vivo Mobile Communication Co., Ltd. \\
$^3$Shenzhen University of Advanced Technology\\
{\tt\small 12333437@mail.sustech.edu.cn \quad \quad pt.jiang@vivo.com }}

\begin{document}
\maketitle
\begin{abstract}
Existing mobile devices are constrained by compact optical designs, such as small apertures, which make it difficult to produce natural, optically realistic bokeh effects. 
Although recent learning-based methods have shown promising results, they still struggle with photos captured under high digital zoom levels, which often suffer from reduced resolution and loss of fine details.
A naive solution is to enhance image quality before applying bokeh rendering, yet this two-stage pipeline reduces efficiency and introduces unnecessary error accumulation.
To overcome these limitations, we propose MagicBokeh, a unified diffusion-based framework designed for high-quality and efficient bokeh rendering.
Through an alternative training strategy and a focus-aware masked attention mechanism, our method jointly optimizes bokeh rendering and super-resolution, substantially improving both controllability and visual fidelity.
Furthermore, we introduce degradation-aware depth module to enable more accurate depth estimation from low-quality inputs.
Experimental results demonstrate that MagicBokeh efficiently produces photorealistic bokeh effects, particularly on real-world low-resolution images, paving the way for future advancements in bokeh rendering.
Our code and models are available at this \href{https://github.com/vivoCameraResearch/MagicBokeh}{url}.
\end{abstract}    
\section{Introduction}
With the rapid advancement of mobile devices, smartphone photography has seen remarkable progress in recent years, which has greatly improved the photo-taking experience for users.
However, limited by hardware constraints, current mobile devices often struggle to produce natural bokeh effects. 
Researchers have proposed many bokeh rendering methods which either rely on physical optics models to simulate light scattering \cite{kraus2007depth, lee2010real, wadhwa2018synthetic, zhang2019synthetic, sheng2024dr} or generate realistic bokeh effects by learning from large-scale datasets \cite{alzayer2023dc2, ignatov2020rendering, peng2022mpib, wang2018deeplens}. 
They can usually generate visually pleasing bokeh results and have been applied on mobile devices.
Despite advances in these methods, one of the major limitations is that they all assume that the input is an all-in-focus high-quality (HQ) image.
When applying these methods to images captured from a high digital zoom of the mobile camera, they often suffer from amplified noise, blurred subject boundaries, and unrealistic texture synthesis.
Moreover, the quality degradation caused by digital zoom in mobile photography further hinders the effectiveness of existing bokeh rendering approaches, where the focused degraded regions usually affect the aesthetics.

\begin{figure}[h]
    \centering
    \includegraphics[width=1\linewidth]{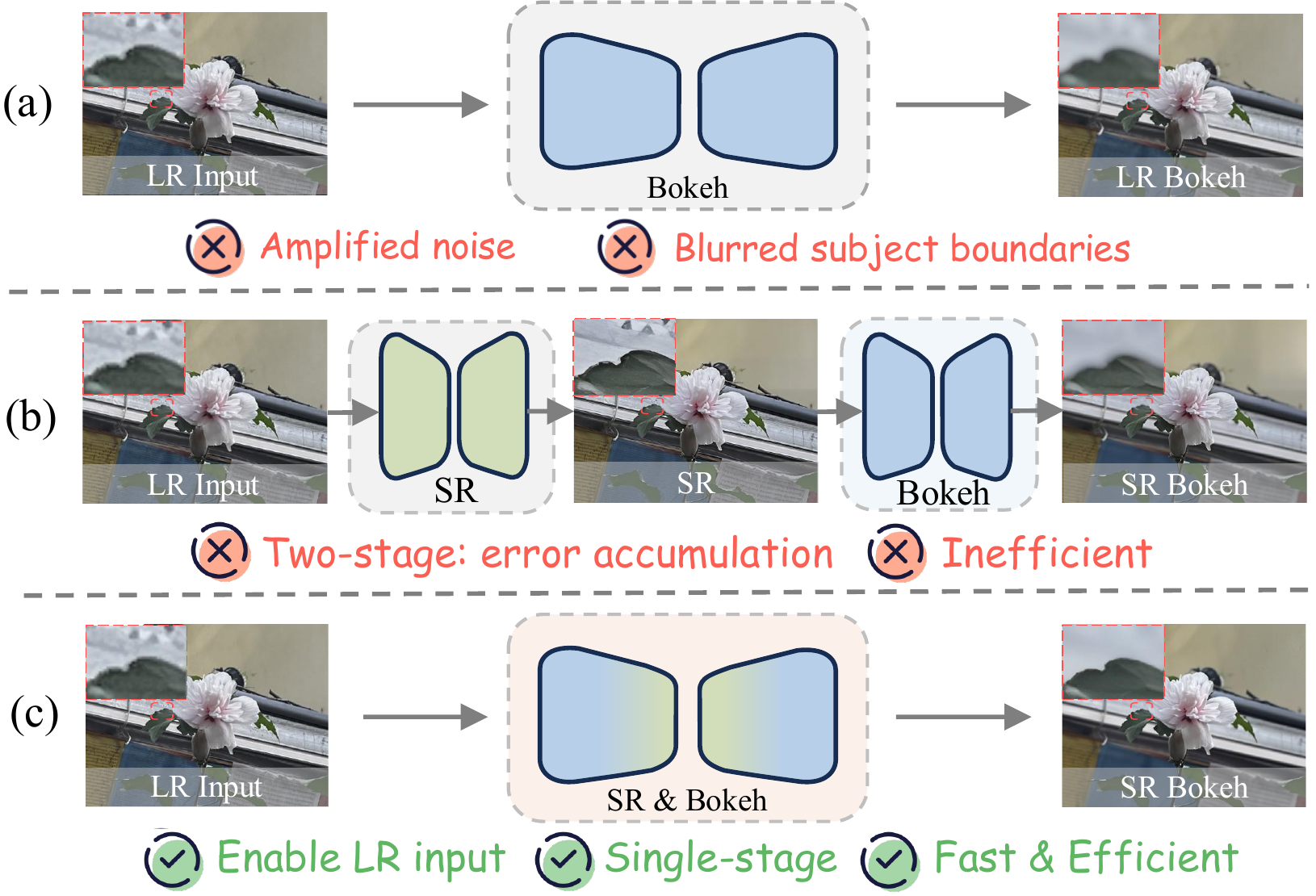}
    \vspace{-20pt}
    \caption{Compared with low-resolution (LR) bokeh rendering (a) and two-stage super-resolution (SR) bokeh rendering (b), our proposed method (c) seamlessly integrates the SR with bokeh rendering within a unified framework, thereby achieving both computational efficiency and photorealistic bokeh effects.}
    \vspace{-15pt}
    \label{fig:motivation}
\end{figure}

To address this issue, a straightforward approach is using a two-stage pipeline:
performing real-world image super-resolution (Real-ISR) first and then conducting bokeh rendering.
However, such a naive approach results in two main problems:
Firstly, since the output of the Real-ISR network is not always perfect, it may introduce error accumulation. 
These errors can be further amplified during the subsequent bokeh rendering process, ultimately degrading the overall image quality, as shown in Fig. \ref{fig:overall}. 
Secondly, the two-stage method requires two separate model inferences, which affects computational efficiency. 
These limitations (shown in Fig. \ref{fig:motivation}) naturally lead us to consider a unified approach.

Recently, diffusion models \cite{ho2020denoising, song2020denoising, song2020score}, such as Stable Diffusion (SD) \cite{rombach2022high}, have demonstrated significant advantages in generating fine-grained image details and show remarkable generalization performance across various tasks, especially in Real-ISR. 
Moreover, we have observed that images produced by generative models often contain inherent bokeh information, indicating that these models possess the bokeh prior. 
This observation motivates us to consider whether we can design a unified diffusion-based approach that improves both the quality and efficiency of bokeh rendering for high digital zoom photography.

In this paper, we present MagicBokeh, a unified diffusion-based single-step framework designed for photorealistic bokeh rendering that can efficiently generate bokeh effects for high-zoom photography.
However, integrating Real-ISR and bokeh rendering into a unified model tends to introduce conflicting optimization objectives between two tasks, leading to performance degradation during training. 
To address this issue, we propose an alternative training strategy and focus-aware mask attention specifically designed for our framework.
To enhance computational efficiency, we compress the computationally intensive U-Net component in SD by block pruning. 
Depth Anything v2 \cite{yang2024depth}, with its powerful depth estimation capability, has been adopted as a depth prior in bokeh rendering tasks. Nevertheless, its performance is still challenged by image quality degradation. To address this issue, we propose a degradation-aware depth module which improves the robustness and accuracy of depth estimation on low-quality (LQ) images.
Experimental results demonstrate that our approach achieves valuable advances in bokeh rendering for high-zoom photography and also performs well in related tasks, such as refocusing. In summary, our main contributions are as follows:
\begin{itemize}
\item We propose MagicBokeh, a diffusion-based single-step framework that conducts Real-ISR and bokeh rendering simultaneously within a unified architecture.
\item To further enhance the quality of image bokeh rendering, we propose an alternative training strategy with focus-aware mask attention and introduce a degradation-aware depth module for improved depth estimation on high-zoom photographs.
\item Comprehensive experiments show that MagicBokeh achieves state-of-the-art (SOTA) quantitative and qualitative results on both synthetic and high-zoom real-world photographs, highlighting its effectiveness in photorealistic bokeh rendering.
\end{itemize}
\section{Related works}
\subsection{Bokeh Rendering}
Bokeh rendering refers to a computational photography technique that simulates the depth-of-field (DoF) effect. 
Existing bokeh rendering methods can be categorized into classical rendering methods and learning-based methods.\\
\textbf{Classical Rendering Methods.} Early bokeh rendering methods were primarily based on classical computer graphics, using ray tracing \cite{pharr2023physically,potmesil1981lens} to generate physically accurate bokeh effects. However, as the camera sampling space increased, the computational complexity increased exponentially, making these methods difficult to render fast. Subsequent methods improve efficiency by providing depth maps and focal plane information \cite{barron2015fast,bertalmio2004real,soler2009fourier,wadhwa2018synthetic,zhang2019synthetic,busam2019sterefo}. 
DeepFocus \cite{senaras2018deepfocus} specializes in using a perfect depth map to render realistic bokeh effects in low resolution. However, obtaining a perfect depth map in the real world is challenging. 
Dr.Bokeh \cite{sheng2024dr} uses an inpainting model to estimate the RGBD values of occluded regions behind the salient object. It then simulates bokeh by computing the scattering and focusing of light in a spherical lens system based on foreground and background images, effectively reducing occlusion artifacts in boundary.
Nevertheless, due to inaccuracies in the disparity maps, these methods often suffer from unnatural partial occlusion artifacts or color bleeding. \\
\textbf{Learning-based Methods.} Recent works~\cite{peng2022bokehme,peng2022mpib,seizinger2023efficient,zhu2025bokehdiff,yang2025any} have introduced neural rendering and generative models to address unnatural partial occlusion artifacts and color bleeding in bokeh rendering. 
BokehMe \cite{peng2022bokehme} first generates bokeh effects using a classical physically motivated renderer and then employs a neural renderer to correct artifacts, mitigating the impact of imperfect disparity inputs. MPIB \cite{peng2022mpib} leverages an inpainting network to restore occluded background regions and applies an adaptive aggregation operation on a multiplane image layer, enabling the network to learn shallow DoF rendering across different focal planes. EBokehNet \cite{seizinger2023efficient} integrates lens properties as additional inputs into the neural network to control the intensity of the bokeh effect. BokehDiff \cite{zhu2025bokehdiff} is a diffusion-based method that achieves accurate results with physics-inspired self-attention. AnytoBokeh \cite{yang2025any} proposes a one-step diffusion framework for temporally coherent, depth-aware video bokeh, leveraging MPI representations and progressive training to achieve stable and controllable blur transitions.
Despite recent advances, these methods still face significant challenges when applied to LQ inputs.

\subsection{Diffusion-based Real-ISR}
Recent advances in generative diffusion models \cite{ho2020denoising,huang2025sdmatte}, particularly large-scale pre-trained text-to-image models \cite{rombach2022high}, have demonstrated exceptional performance in various downstream tasks, especially in ISR tasks \cite{lin2024diffbir,xie2024addsr,moser2024diffusion,yu2024scaling}. Recent studies have increasingly focused on single-step diffusion ISR models \cite{wang2024sinsr,wu2024one,zhang2024degradation}, which have shown great value when used on mobile devices. 
SinSR \cite{wang2024sinsr} presents a deterministic sampling technique that stabilizes the noise-image pair through consistency-preserving distillation. OSEDiff \cite{wu2024one} employs variational score distillation \cite{wang2023prolificdreamer} to maintain fidelity when generating high-resolution images. S3Diff \cite{zhang2024degradation} leverages the T2I prior from SD-Turbo \cite{sauer2024adversarial} to achieve HQ images in a single step. 
Reference-based Ada-RefSR \cite{wang2026trust} adaptively regulates reference guidance to mitigate hallucinations. 
RCOD \cite{wu2026realism} introduces latent-domain grouping and degradation-aware sampling to flexibly control fidelity–realism trade-offs.
Inspired by the aforementioned methods, we integrate the single-step Real-ISR task into the bokeh rendering pipeline to enhance both generation quality and efficiency.
\section{Methodology}

\begin{figure*}
  \centering
  \includegraphics[width=\linewidth]{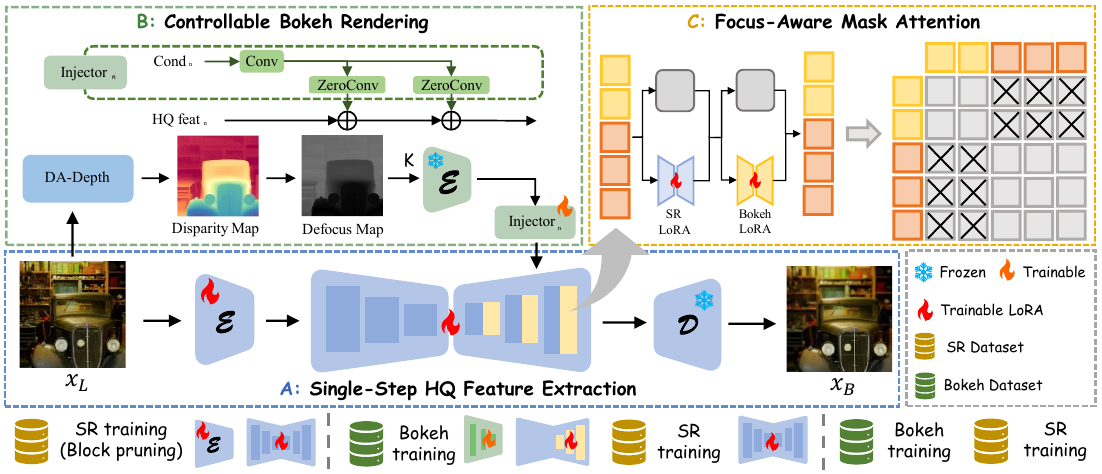}
  \vspace{-0.6cm}
  \caption{The framework of MagicBokeh. We introduce an alternative training strategy to unified Real-ISR and bokeh rendering together. During the bokeh training, the Controlnet and bokeh LoRA layers are trainable to learn controllable bokeh rendering. During the Real-ISR training, only the SR LoRA is trainable to learn SR. During inference, given a high-zoom LQ photo, it can generate a disparity map through the degradation-aware depth model to guide bokeh rendering.}
  \label{fig:2}
  \vspace{-0.6cm}
\end{figure*}

\subsection{Framework Overview}\label{sec:4.1}
Existing bokeh rendering methods based on generation models or lens blur rendering often rely on HQ image input, which is not suitable for high-zoom mobile photography. 
Therefore, we propose MagicBokeh, a diffusion-based framework that is highly suitable for this task while maintaining computational efficiency.
As illustrated in Fig. \ref{fig:2}, MagicBokeh consists of two main parts: HQ feature extraction and controllable bokeh rendering. 
The former extracts HQ features from LQ images, while the latter governs bokeh rendering based on the controllable bokeh rendering module and focus-aware mask attention. \\
\textbf{Single-Step HQ Feature Extraction.} 
Recent diffusion-based ISR approaches \citep{yue2024difface, li2023dissecting} have shown that directly using LQ images with little or no noise as input can substantially eliminate the uncertainty introduced by random noise sampling, while maximizing the retention of semantic content. Therefore, we directly feed the LQ images into the HQ feature extraction module without introducing any noise. Then, we inject Low-Rank Adaptation (LoRA) \citep{hu2022lora} into both the VAE encoder (only train in the first ISR training stage) and modified lightweight U-Net, and finetune the model to recover its HQ feature extraction capability. We employ L2 loss and LPIPS loss for supervision.
\\
\textbf{Controllable Bokeh Rendering Module.}
To achieve precise and controllable bokeh rendering, we introduce ControlNet \citep{zhang2023adding} as a conditional control module. 
In our framework, ControlNet receives a defocus map as the structural condition. Specifically, we first estimate a disparity map from the depth estimation network. 
The defocus map can be calculated by
\begin{equation}
r = K \left| d - d_f \right|, 
\end{equation}
where $d$ represents the disparity of the pixel, $d_f$ denotes the disparity of the focal position that the users specified, $K$ indicates the blur intensity, and $r$ represents the blur radius of the pixel.
By integrating ControlNet, our model can generate visually plausible bokeh with controllable depth-of-field (DoF), while preserving semantic consistency in the in-focus regions.\\
\vspace{-5pt}
\subsection{Alternative Training Strategy}
When implementing end-to-end training of our MagicBokeh framework using the SR bokeh dataset (containing paired LQ and HQ bokeh images, mentioned in Sec. \ref{SRBOKEHDATASET}), we observed a notable performance degradation in the ISR of subject areas, despite the original intention to simultaneously optimize both subject super resolution and background bokeh rendering, as shown in the top part of Fig. \ref{fig:8}.
This degradation primarily arises from the conflicting optimization objectives inherent in Real-ISR and bokeh rendering tasks. 
Furthermore, the imbalance between the training samples for these tasks biases the network toward optimizing one task at the expense of the other.

To effectively address these challenges and mitigate conflict between tasks, we propose an alternative training strategy to decouple Real-ISR from bokeh rendering.  
This cyclical strategy alternates attention between different tasks.
Before applying the alternating training strategy, the model is first initialized through super-resolution pre-training. This gives the network strong Real-ISR capability and a better starting point for subsequent joint optimization.
In bokeh rendering phase, training emphasizes HQ bokeh rendering using LQ all-in-focus images as inputs, conditioned by defocus maps to generate HQ bokeh outputs.
During this stage, the original diffusion model and pre-trained HQ feature extraction model are fixed, and training specifically targets the ControlNet and the bokeh LoRA layers in the focus-aware mask attention modules to refine the quality of bokeh rendering.
Subsequently, training shifts to Real-ISR, employing pairs of LQ and HQ images as training samples. 
In this phase, a defocus map with all-zero values is used as input to represent an all-in-focus condition, while the optimization is restricted solely to the SR LoRA layers within the UNet of the diffusion network.
We alternatively train these two phases.
Our experiments validate that by alternating the focus between bokeh rendering and Real-ISR tasks, our proposed training strategy effectively reduces intertask interference, ultimately achieving significant improvements in the quality of bokeh rendering.

\subsection{Focus-aware Mask Attention}
In our task, incorporating bokeh conditions directly into the generation process frequently results in degradation of the restoration quality for focused regions.
To address this issue, we propose an approach that explicitly decouples Real-ISR from bokeh rendering, ensuring that the in-focus areas are accurately reconstructed without being affected by the defocused regions.
Notably, in text-to-image models such as SD, self-attention layers play a crucial role in maintaining global coherence within generated images. Previous research \citep{epstein2023diffusion,kim2023dense} has shown that appropriately modulating self-attention layers can significantly enhance the controllability of generative results.

Although employing the alternative training strategy can alleviate conflicts between these two tasks, incorrect control still persists, particularly in image details. Consequently, we propose focus-aware mask attention, as shown in Fig. \ref{fig:2}c, which utilizes focus cues obtained through the defocus maps as guidance for modulating self-attention layers. 
Specifically, we modulate the attention maps as below,
\begin{equation}
\text{Attention} = \text{softmax}\left(\frac{\mathbf{Q}\mathbf{K}^{\top} + \mathcal{M}}{\sqrt{d}}\right) \mathbf{V},
\end{equation}
where $\mathbf{Q}$, $\mathbf{K}$, $\mathbf{V}$ are the query, key and value of the self-attention layer, respectively.
The focus attention mask $\mathcal{M}$ at feature location $(x, y)$ is
\begin{equation}
\mathcal{M}_{(x, y)} = 
\begin{cases} 
0 & \text{if } \mathbf{M}_{(x, y)} = 1 \\
-\infty & \text{otherwise}
\end{cases}, 
\end{equation}
where $\mathbf{M}$ is the binary result obtained by extracting the subject information from the defocus map in the focus region and binarizing the relationships between different regions (with the same regions being 1 and different regions being 0). This binary mask is resized to match the resolution required by the attention layer.
During the training process, we alternately trained the SR LoRA layer and Bokeh LoRA layer.
Notice that in the Real-ISR phase, the attention mask $\mathcal{M}$ is set to 0 to restore the whole image. 

Integrating this attention mechanism into the proposed alternative training strategy enables a clear delineation of tasks: specifically, by effectively partitioning the disparity map into binary foreground–background regions and constraining self-attention to operate within each region, the ISR component is guided to prioritize the focused subject area, while the bokeh rendering component is steered toward enhancing background bokeh effects.
Our experimental results demonstrate that the focus-aware mask attention substantially enhances the controllability of our unified model, thus improving the quality of generated images.
\subsection{Degradation-aware Depth Estimation}
Despite the remarkable performance in the HQ data, the accuracy of the depth estimation model deteriorates rapidly when applied to LQ images. The input of imperfect disparity map degrades the results of the SR bokeh rendering. To address this issue, we propose a self-feature distillation framework to estimate HQ-like features. We utilize the pre-trained Depth Anything v2 \cite{yang2024depth} as the baseline network for both the teacher and student models. During the training process, both HQ images and simulated degraded images are respectively input into the teacher and student networks to extract features from encoder. Through feature distillation and output supervision, the features are expected to remain consistent, thereby improving the performance of depth estimation. Additional analyses of the results are provided in the supplementary material.
\section{Experiment}
\subsection{Experimental Setup}
\textbf{Training Datasets. \label{SRBOKEHDATASET}} Following the setup of recent works \citep{wu2024seesr, wu2024one}, we train our HQ feature extraction model on the LSDIR \citep{li2023lsdir} and a subset of 10k face images from FFHQ \citep{karras2019style}. Additionally, to obtain HQ bokeh images as ground truth in bokeh training stage, similar to MPIB \citep{peng2022mpib} and Dr.Bokeh \citep{sheng2024dr}, we built a ray-tracing-based renderer that generates lens blur through a real thin lens. More details are provided in the supplementary material.
During the training process, we use the degradation pipeline proposed in Real-ESRGAN \citep{wang2021real} to synthesize the required LQ-HQ pairs. 
The synthesized LQ images are upscaled to match the HR resolution of 512 $\times$ 512 before feeding into our model.\\
\textbf{Evaluation Metrics.}
We evaluate the performance of various methods using both full-reference and no-reference metrics. First, we use PSNR, SSIM \citep{wang2004image}, and LPIPS \citep{zhang2018unreasonable} to measure the fidelity of the bokeh rendering. We also use reference-based perceptual metrics such as DISTS \citep{ding2020image}, image generation similarity metrics like FID \citep{heusel2017gans}, and no-reference metrics including NIQE \citep{zhang2015feature}, MANIQA \citep{yang2022maniqa}, MUSIQ \citep{ke2021musiq}, and CLIPIQA \citep{wang2023exploring}.\\
\textbf{Implementation Details.}
Our single-step HQ feature extraction model is built upon SD2.1, where we remove all cross-attention layers and the mid-stage module in the original U-Net by block pruning.
Specifically, through experimental observations on existing single-step Real-ISR methods \citep{wu2024one, zhang2024degradation}, we observe that while text prompts provide semantic information, they offer limited benefits and significant computational overhead in extracting HQ features in practice.  
Consequently, we remove the text encoder and cross-attention modules from the pipeline, effectively eliminating prompt dependency and reducing computational overhead. 
Following \citep{kim2023bk}, we streamline the U-Net architecture by removing the entire mid-stage module, which significantly improves efficiency without compromising perceptual quality.
We inject LoRA modules into both the VAE encoder and the modified lightweight U-Net, and retrain the model on the Real-ISR dataset with paired LQ-HQ images using the AdamW optimizer with a learning rate of 5$e$-5.
Then, we adopt an alternative training strategy consisting of two phases.  
In the bokeh rendering phase, we train the controllable bokeh rendering module and the bokeh LoRA layers in the focus-aware mask attention module on the SR bokeh dataset containing paired LQ and HQ bokeh images. 
The learning rate is set to 5$e$-5.
In the following Real-ISR phase, we train the SR LoRA layers of the UNet on the ISR dataset, employing a learning rate of 5$e$-6.
The entire training process takes approximately 20 hours on 4 NVIDIA L40 GPUs. 
In addition, we apply random horizontal flipping to enhance the diversity of training data.

\subsection{Results on Synthetic Degradation Dataset}
\textbf{Synthetic Degradation Dataset.} We conduct a systematic evaluation of bokeh rendering performance on the real-world EBB dataset.
\begin{figure*}[t]
  \centering
  \includegraphics[width=\linewidth]{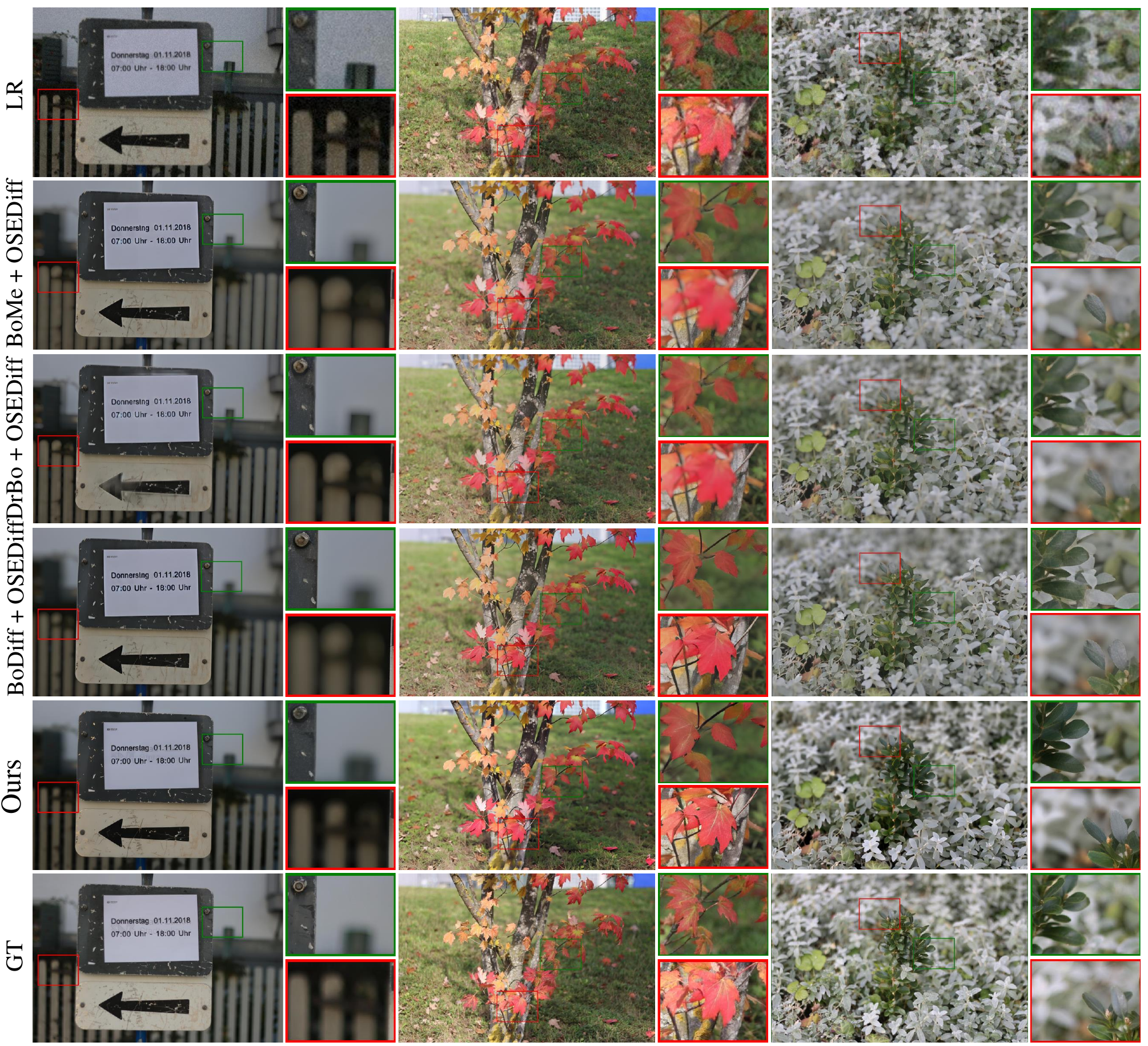}
  \vspace{-0.6cm}
  \caption{Qualitative comparison on EBB400-LQ. More results can be seen in the supplementary material.}
  \label{Fig:6}
  \vspace{-0.6cm}
\end{figure*}
For the established EBB400 benchmark, we randomly select 400 image pairs and manually label the focal regions in each image to assess the bokeh rendering accuracy. 
We applied this benchmark to evaluate images in the high-zoom bokeh rendering task, named EBB400-LQ, where image degradation was simulated using the Real-ESRGAN pipeline.
To ensure a fair comparison, since the compared methods obtain SR images after the first stage, Depth Anything v2 \cite{yang2024depth} is used to generate disparity maps. In contrast, our approach employs the proposed degradation-aware depth module to estimate more robust disparity maps from the original LQ inputs.
All disparity maps are normalized to 0-1 during testing.\\
\noindent\textbf{Experimental Results.}
To validate the effectiveness of our method, we compare MagicBokeh with two-stage pipelines, including SOTA diffusion-based Real-ISR methods and bokeh rendering methods. 
Specifically, considering that recent works have focused mainly on the diffusion-based single-step framework, we evaluate our method against Real-ISR methods including OSEDiff \citep{wu2024one}, and S3Diff \citep{zhang2024degradation}. 
The approaches which require depth maps always get better bokeh effects, so we compare with bokeh rendering methods including BokehMe \citep{peng2022bokehme}, Dr.Bokeh \citep{sheng2024dr} and BokehDiff \citep{zhu2025bokehdiff}. 
As shown in the Tab. \ref{table:1}, our model achieves SOTA performance compared to previous two-stage SOTA methods, demonstrating its superior effectiveness in high digital zoom bokeh rendering. 
\begin{table*}[t]
\centering
\caption{Quantitative comparison of performance with two-stage SOTA models on EBB400-LQ benchmark. ISR methods use OSEDiff (*) and S3Diff($^{+}$). The inference times are tested with an input image of size 512 × 512, and the inference time is measured on an L40s GPU. \textbf{Bold} and \underline{underline} denote the best and the second best result.}
\vspace{-0.2cm}
\label{table:1}
\resizebox{\linewidth}{!}{
\begin{tabular}{c|cccccccc}
\toprule[1.0pt]
Method & PSNR $\uparrow$ & SSIM $\uparrow$ & LPIPS $\downarrow$ & DISTS $\downarrow$ & MUSIQ $\uparrow$ & MANIQA $\uparrow$ & FID $\downarrow$ & Time(s) $\downarrow$\\
\midrule
    BokehMe* & 23.51 & 0.8459 & 0.3106 & 0.1666 & 57.70 & 0.4219 & 72.98 & \underline{0.1648} \\
    Dr.Bokeh* & 23.39 & \underline{0.8488} & 0.3132 & 0.1677 & 52.40 & 0.3934 & 73.38 & 2.4021 \\
    BokehDiff* & 23.65 & 0.8459 & \underline{0.3049} & 0.1713 & \underline{59.24} & \underline{0.4251} & 72.65 & 0.3376 \\
    BokehMe$^{+}$ & 23.75 & 0.8388 & 0.3138 & \underline{0.1606} & 57.54 & 0.4137 & \textbf{72.25} & 0.7510 \\
    Dr.Bokeh$^{+}$ & 23.67 & 0.8430 & 0.3134 & 0.1687 & 52.63 & 0.3876 & 73.10 & 2.9883 \\
    BokehDiff$^{+}$ & \underline{23.83} & 0.8397 & 0.3071 & 0.1735 & \textbf{59.36} & \textbf{0.4259} & 72.54 & 0.9238 \\
    MagicBokeh & \textbf{24.23} & \textbf{0.8623} & \textbf{0.2786} & \textbf{0.1600} & 58.83 & 0.4138 & \underline{72.43} & \textbf{0.1062} \\
\bottomrule[1.0pt]
\end{tabular}
}
\vspace{-0.4cm}
\end{table*}
Although our method performs worse than BokehDiff in some non-parameterized metrics, this is mainly because BokehDiff produces inaccurate focus distributions in the EBB400-LQ dataset, where regions that should exhibit bokeh remain in focus, leading to higher metric values. However, these gains do not reflect realistic bokeh effects. As shown in Fig. \ref{Fig:6} and in the supplementary material, our method produces more visually plausible results. The comparisons also highlight clear limitations of existing two-stage methods.
Firstly, the two-stage methods require two separate model inferences, which lead to inefficiency.
Second, these methods fail to produce realistic bokeh effects in complex natural scenes, as shown in the third example of Fig. \ref{Fig:6}.
Moreover, the edge artifacts introduced during Real-ISR lead to bokeh rendering with unnatural edge transitions, as shown in the second example of Fig. \ref{Fig:6}.
In contrast, by reusing the prior information from the diffusion model and adopting an alternative training strategy along with focus-aware mask attention, our approach delivers superior bokeh quality and computational efficiency compared to other methods. 
Furthermore, although we do not use any text conditions, MagicBokeh still shows strong performance in the task of Real-ISR compared with single-step Real-ISR methods, as illustrated in the supplementary material, highlighting its ability to restore both the realism and aesthetic quality of images. 

\subsection{User study on Real-world Dataset}
\textbf{Real-world Degradation Dataset.} Synthetic degradation datasets fail to capture the complex artifacts in real-world photography, such as hybrid sensor circuit noise, motion blur from handheld shooting and lossy compression in digital zoom. To address this, we design a user study specifically on authentic LQ images captured under practical mobile photography conditions. We collected 50 real-world LQ images using an iPhone 13 pro, covering diverse scenarios (portraits, landscapes, indoor/outdoor scenes) with varying high digital zoom levels (5$\times$ – 15$\times$). The average resolution of the images is 4032 $\times$ 3024.

\begin{figure}[h]
  \centering
  \includegraphics[width=0.8\linewidth]{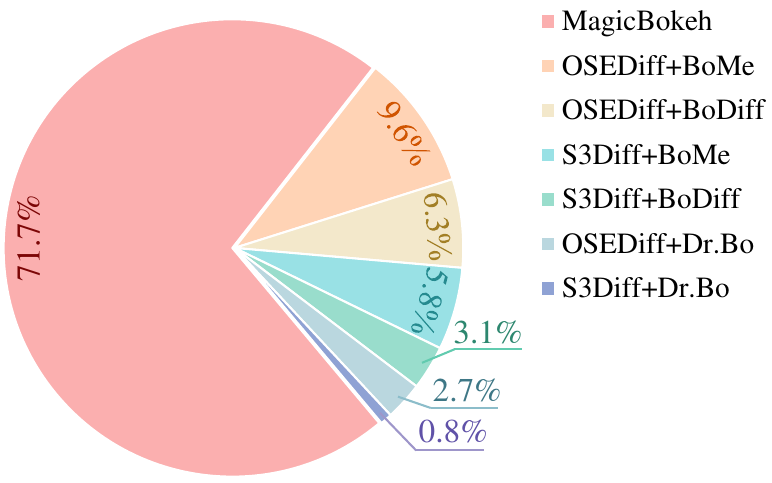}
  \caption{The human preference on the real-world results.}
  \vspace{-12pt}
  \label{fig:7}
\end{figure}

\noindent \textbf{Quantitative Results.} 
This study engages 50 participants from diverse backgrounds, ensuring a wide range of perspectives. Each participant is presented with bokeh images from different methods, and they are then asked to choose the best one from these images. As shown in Fig. \ref{fig:7}, our method achieves outstanding scores compare with other two-stage approaches in the HQ bokeh rendering task for high-zoom mobile photography.

\subsection{Ablation Studies}
In this section, we perform a comprehensive ablation study to assess the impact of each component in MagicBokeh on the EBB400-LQ dataset. 

\begin{table*}[h]
  \centering
  \caption{Ablation study on the EBB400-LQ dataset. The setting of “FAMA”, “Strategy”, and “DA depth” are short for the focus-aware mask attention, alternate training strategy, and degradation-aware depth module respectively. \textbf{Bold} and \underline{underline} denote the best and the second best result.}
  \vspace{-5pt}
  \label{table:3}
    \resizebox{\textwidth}{!}{
      \begin{tabular}{cccccccccc}
        \toprule[1.0pt]
        \multicolumn{3}{c}{Modules} & \multicolumn{7}{c}{Metrics}\\
        \midrule 
        FAMA & Strategy & DA depth &PSNR $\uparrow$ & LPIPS $\downarrow$ & CLIP-IQA $\uparrow$ & NIQE $\downarrow$ & MUSIQ $\uparrow$ & MANIQA $\uparrow$ & FID $\downarrow$\\
        \midrule
         \XSolidBrush & \XSolidBrush & \XSolidBrush & 24.21 & 0.2931 & 0.3743 & 6.0786 & 57.41 & 0.4038 & 73.25 \\
         \XSolidBrush & \Checkmark & \Checkmark & \underline{24.22} & 0.2798 & 0.4157 & 5.9068 & 58.10 & 0.4065 & 75.23 \\
         \Checkmark & \XSolidBrush & \Checkmark & 24.20 & 0.2946 & 0.3781 & \underline{5.7076} & 56.08 & 0.3956 & \underline{73.04} \\
         \Checkmark & \Checkmark & \XSolidBrush & 24.20 & \textbf{0.2784} & \underline{0.4209} & 5.8035 & \underline{58.80} & \underline{0.4114} & 75.03 \\
         \Checkmark & \Checkmark & \Checkmark & \textbf{24.23} & \underline{0.2786} & \textbf{0.4229} & \textbf{5.6341} & \textbf{58.83} & \textbf{0.4138} & \textbf{72.43} \\
        \bottomrule[1.0pt]
    \end{tabular}
    }
\end{table*}
\begin{figure*}
  \centering
  \includegraphics[width=\linewidth]{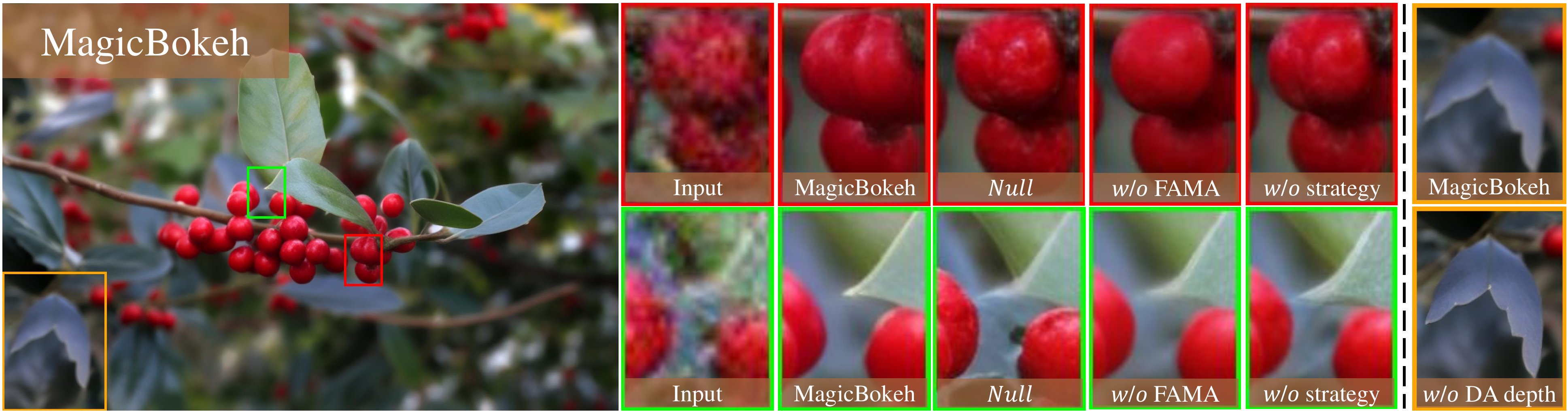}
  \caption{Visual comparison of the ablation study.}
  \vspace{-10pt}
  \label{fig:8}
\end{figure*}

\noindent \textbf{Effect of Focus-aware Mask Attention.} To validate whether focus-aware mask attention can effectively reconstruct the focal region while being unaffected by the defocused areas, we designed a single-contrast variant model, referred to as \textit{w/o} focus-aware mask attention (\textit{w/o} FAMA). This variant model does not use the focal cues obtained from the defocused image to modulate self-attention, instead applying attention operations only on the global image. The results are shown in Tab. \ref{table:3}. As can be seen, the full model and \textit{w/o} FAMA seems minor differences in PSNR, the no-reference metrics show significant improvement in the full model. This indicates that the focus-aware mask attention mechanism can successfully decouple the focused subject from the out-of-focus area.

\noindent \textbf{Effect of Alternate Training Strategy.} To verify whether the alternate training strategy improves the quality of subject Real-ISR and the blurring effect of the defocus region, we designed another single contrast variant model, named \textit{w/o} alternative training strategy (\textit{w/o} Strategy). The results are shown in Tab. \ref{table:3} and Fig. \ref{fig:8}. As can be seen, compared to \textit{w/o} strategy, the full model, which includes the alternative training strategy, enhancing the quality of bokeh rendering. 

\noindent \textbf{Effect of Degradation-Aware Depth Module.} To assess the contribution of degradation-aware depth module in MagicBokeh, we input the disparity map predicted by Depth Anything v2 \cite{yang2024depth} into the network and conduct a comparative experiment, named \textit{w/o} DA depth. 
To verify, the results are listed in Tab. \ref{table:3}. Although there is no substantial difference in quantitative metrics between \textit{w/o} DA depth and our method, we can find improvement in qualitative comparison, as shown in Fig. \ref{fig:8}. DA depth provides better depth estimation results for LQ images.

\begin{figure}[h]
  \centering
  \vspace{-0.2cm}
  \includegraphics[width=\linewidth]{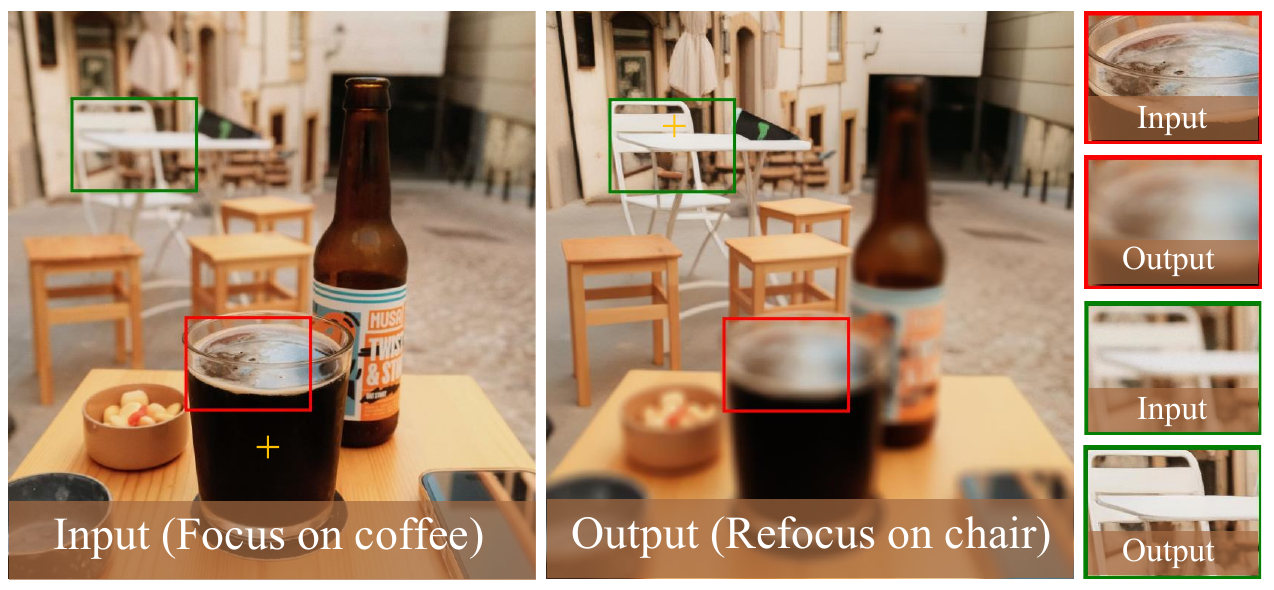}
  \caption{Further application in refocusing.}
  \vspace{-0.2cm}
  \label{Fig:8}
\end{figure}

\subsection{Further Application}\label{Further Application}
While existing bokeh rendering methods assume all-in-focus inputs, photographs often contain partially defocused regions due to autofocus errors or multi-subject compositions. Thus, reconstructing sharp image areas that are blurred by the bokeh effect and refocusing on new regions of interest presents a critical challenge.
Our method, which is built upon LQ input images, is found to generalize well to the task of refocusing. As shown in Fig. \ref{Fig:8}, the result demonstrates that our approach significantly produces smooth blur transitions when shifting focus from the coffee cup to background chairs.
\section{Conclusion}
\vfill
In this paper, we present MagicBokeh, a unified diffusion-based framework designed for photorealistic and efficient bokeh rendering. Our method jointly performs Real-ISR and bokeh rendering in the unified architecture, thereby effectively overcoming the limitations of traditional two-stage pipelines. To address the conflicting objectives between Real-ISR and bokeh rendering, we introduce an alternating training strategy that enables the model to learn both tasks efficiently. Furthermore, we design two plug-and-play modules, namely controllable bokeh rendering and focus-aware mask attention, to guide bokeh rendering and enhance subject-background separation, respectively. Extensive experiments demonstrate that MagicBokeh achieves SOTA results in high-zoom bokeh rendering and is well suited for robust real-world photography applications.

\clearpage
\renewcommand{\thesection}{\Alph{section}}
\setcounter{section}{0}

\setcounter{figure}{0} 
\renewcommand{\thefigure}{s\arabic{figure}}

\setcounter{table}{0} 
\renewcommand{\thetable}{s\arabic{table}}

\section{Degradation-aware Depth Estimation}
\subsection{Training Details}
Despite the remarkable performance in HQ data, the accuracy of the depth estimation model deteriorates rapidly when applied to LQ images. And the input of imperfect disparity map degrades the results of SR bokeh rendering. 

\begin{figure}[h]
  \centering
  \vspace{-0.3cm}
  \includegraphics[width=0.9\linewidth]{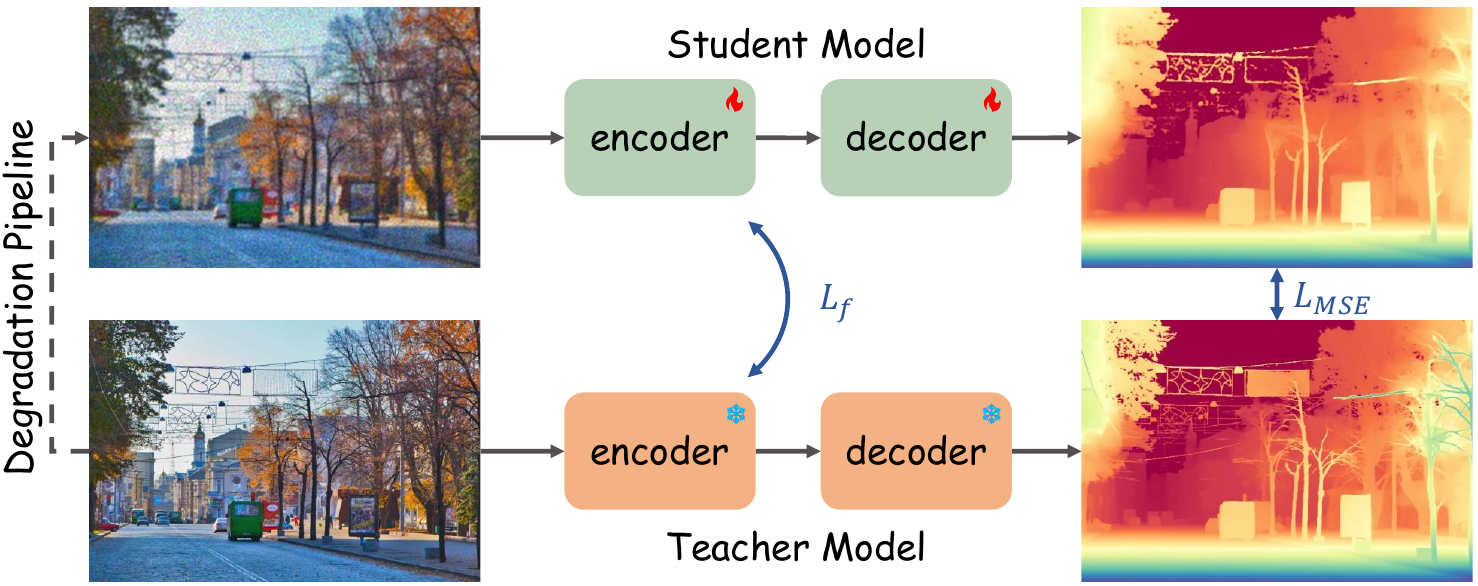}
  \vspace{-0.2cm}
  \caption{The training pipeline of the DA depth module.}
  \vspace{-0.6cm}
  \label{fig:4}
\end{figure}

To address this issue, we propose a self-feature distillation framework to estimate HQ-like features. As shown in Fig. \ref{fig:4}, we utilize the pre-trained Depth Anything v2 Large model as the baseline network for both the teacher and the student models. During the training process, both HQ images and simulated degraded images are respectively input into the teacher and student networks to extract features from encoder. Through feature distillation, features are expected to remain consistent, thereby improving depth estimation performance. Simultaneously, the network's output is supervised to obtain a more accurate depth map.

\subsection{Quantitative comparison of depth estimation}
In our experiments, we use the pre-trained Depth Anything v2 as the teacher model to generate pseudo-labels and supervise the student model, initialized identically, within a distillation framework that takes only RGB images as input. 
Specifically, we conduct our distillation experiments using a subset of 200,000 samples from the SA-1B dataset \cite{kirillov2023segment}. The Real-ESRGAN degradation pipeline \cite{wang2021real} is used to synthesize LQ-HQ training pairs. 

\begin{table*}[h]
  \centering
  \caption{Quantitative comparison on the NYUv2 and KITTI datasets (seen datasets with synthetic degradations) for “Degrade”, “Clear”, and “Average” scenarios.}
  \vspace{-0.2cm}
  \label{table:2}
  \resizebox{0.7\textwidth}{!}{
    \begin{tabular}{cccccccc}
      \toprule
      \multirow{2}{*}{\textbf{Dataset}} & \multirow{2}{*}{\textbf{Method}} & \multicolumn{2}{c}{\textbf{Degrade}} & \multicolumn{2}{c}{\textbf{Clear}} & \multicolumn{2}{c}{\textbf{Average}} \\
      \cmidrule(lr){3-4}\cmidrule(lr){5-6}\cmidrule(lr){7-8}
      & & AbsRel $\downarrow$ & $\delta_{1}$$\uparrow$ & AbsRel$\downarrow$ & $\delta_{1}$$\uparrow$ & AbsRel$\downarrow$ & $\delta_{1}$$\uparrow$ \\
      \midrule
      \multirow{2}{*}{NYUv2} & Depth Anything v2 & 0.081 & 0.926 & \textbf{0.043} & \textbf{0.981} & 0.062 & 0.954 \\
      & DA depth        & \textbf{0.068} & \textbf{0.946} & 0.047 & 0.976 & \textbf{0.058} & \textbf{0.961} \\
      \midrule
      \multirow{2}{*}{KITTI} & Depth Anything v2 & 0.123 & 0.852 & \textbf{0.074} & \textbf{0.946} & 0.099 & 0.899 \\
      & DA depth        & \textbf{0.105} & \textbf{0.883} & 0.079 & 0.944 & \textbf{0.092} & \textbf{0.914} \\
      \bottomrule
    \end{tabular}
    }
    \vspace{-0.2cm}
\end{table*}

To demonstrate the effectiveness of our degradation-aware depth model on degraded images, we compare our approach with Depth Anything v2. 
Tab. \ref{table:2} shows that our method outperforms these related works on the degraded NYUv2 \cite{silberman2012indoor} and KITTI \cite{geiger2013vision}. 
We use point prompts for "Degrade", "Clear" and "Average" scenarios. "Degrade" refers to images degraded by Real-ESRGAN, "Clear" refers to the original, non-degraded images, and "Average" is the mean value of the "Degrade" and "Clear" images. Through self-feature distillation, our student model not only exhibits minimal performance degradation on clear images but also outperforms the baseline on degraded images, thereby verifying the superiority of our method. 

\section{Detail of bokeh training datasets\label{Appendix.2}}
\begin{figure}[h]
  \centering
  \vspace{-0.4cm}
  \includegraphics[width=0.83\linewidth]{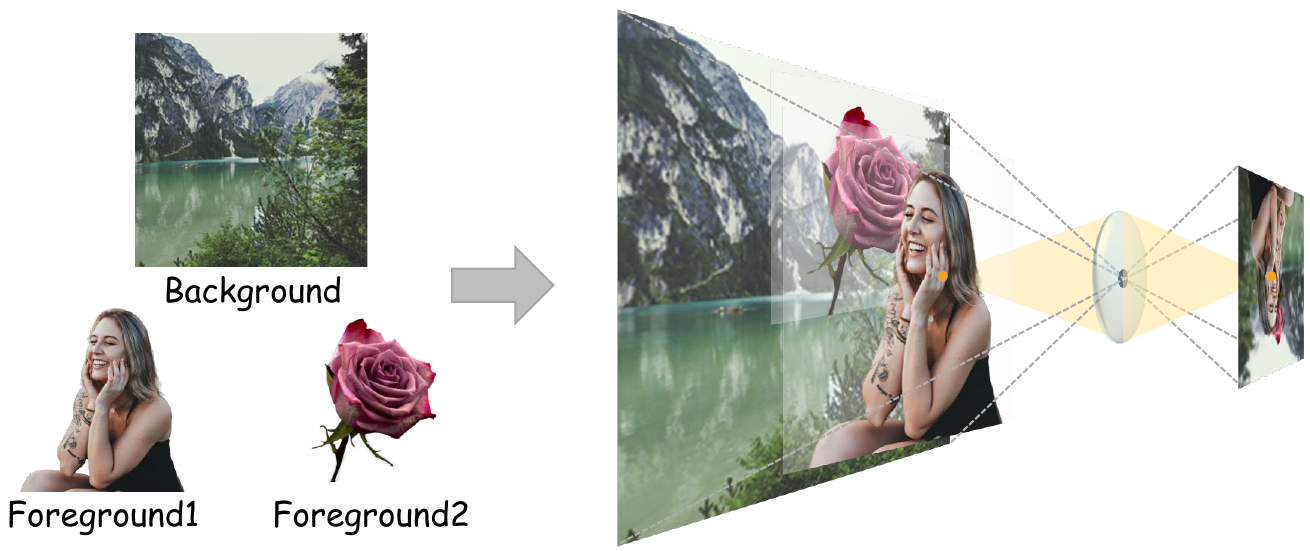}
  \vspace{-0.2cm}
  \caption{The pipeline of data synthesis.}
  \vspace{-0.5cm}
  \label{dataset}
\end{figure}

To obtain HQ bokeh images as ground truth in bokeh training stage, similar to MPIB \cite{peng2022mpib} and Dr.Bokeh \cite{sheng2024dr}, we built a ray-tracing-based renderer that generates lens blur through a real thin lens, as shown in Fig. \ref{dataset}. We first collected nearly 2k high-resolution landscape images from the Internet to serve as our background images. The foreground images are collected from PhotoMatte85 \cite{lin2021real}, RWP-636 \cite{yu2021mask}, AIM-500 \cite{li2021deep} and websites. Each sample is randomly composed of two selected foreground images and one background image. 
During the composition process, the disparity map is set within the range from 0 to 1, the random blur parameter ranges from 0 to 32, and the disparity focus is randomly set to one of the positions in either the foreground or the background. In order to introduce more variation in depth and create more diverse blur effects in the training data, we randomly set the depth variation for the background.

\begin{table*}[t]
    \caption{Quantitative comparison with state-of-the-art methods on real-world benchmarks (RealSR \cite{cai2019toward} and DrealSR \cite{wei2020component}). By providing a defocus map with all-zero input, our method can generate a high-quality all-in-focus image for quantitative comparison. The best and second-best results are highlighted in \textbf{bold} and \underline{underline}.}
    \vspace{-0.2cm}
    \label{table:5}
    \resizebox{1\linewidth}{!}{
    \begin{tabular}{ccccccccccl}
        \toprule
        \textbf{Datasets} & \textbf{Methods} & \textbf{PSNR $\uparrow$} & \textbf{SSIM $\uparrow$} & \textbf{LPIPS $\downarrow$} & \textbf{DISTS $\downarrow$} & \textbf{CLIP-IQA $\uparrow$} & \textbf{NIQE $\downarrow$} & \textbf{MUSIQ $\uparrow$} & \textbf{MANIQA $\uparrow$} & \textbf{FID $\downarrow$} \\
        \midrule{\multirow{4}{*}{RealSR}}
        & SinSR & \textbf{26.32} &  \underline{0.7363} & 0.3195 & 0.2351 & 0.6153 & 6.3541 & 60.42 & 0.5366 & 138.64 \\
        & OSEDiff & 25.15 & 0.7341 & 0.2921 & \underline{0.2128} &  \underline{0.6685} & 5.6528 &  \textbf{69.11} &  \underline{0.6332} & 123.68 \\
        & S3Diff & 25.18 & 0.7269 & \textbf{0.2722} & \textbf{0.2005} &  \textbf{0.6742} & \textbf{5.2612} &  \underline{67.82} &  \textbf{0.6417} & \textbf{105.11} \\
        & MagicBokeh &  \underline{26.14} & \textbf{0.7392} & \underline{0.2888} & 0.2192 & 0.6246 & \underline{5.6337} & 67.22 & 0.6214 & \underline{123.06} \\
        \midrule{\multirow{4}{*}{DRealSR}}
        & SinSR & \underline{28.35} & 0.7484 & 0.3689 & 0.2497 & 0.6319 & 6.9533 & 55.09 & 0.4881 & 170.18 \\
        & OSEDiff & 27.91 & \underline{0.7834} & \textbf{0.2968} & 0.2269 &  \underline{0.6964} & 6.4907 &  \textbf{64.65} & 0.5899 & \underline{135.28} \\
        & S3Diff & 27.54 & 0.7491 & 0.3109 & \textbf{0.2100} &  \textbf{0.7132} & \underline{6.1935} &  \underline{63.93} &  \textbf{0.6099} & \textbf{118.57} \\
        & MagicBokeh &  \textbf{28.99} & \textbf{0.7901} & \underline{0.3003} & \underline{0.2220} & 0.6633 & \textbf{6.1204} & 62.93 &  \underline{0.5901} & 143.02 \\
        \bottomrule
    \end{tabular}
    }
    \vspace{-0.2cm}
\end{table*}

\begin{figure*}[!ht]
  \centering
  \includegraphics[width=\linewidth]{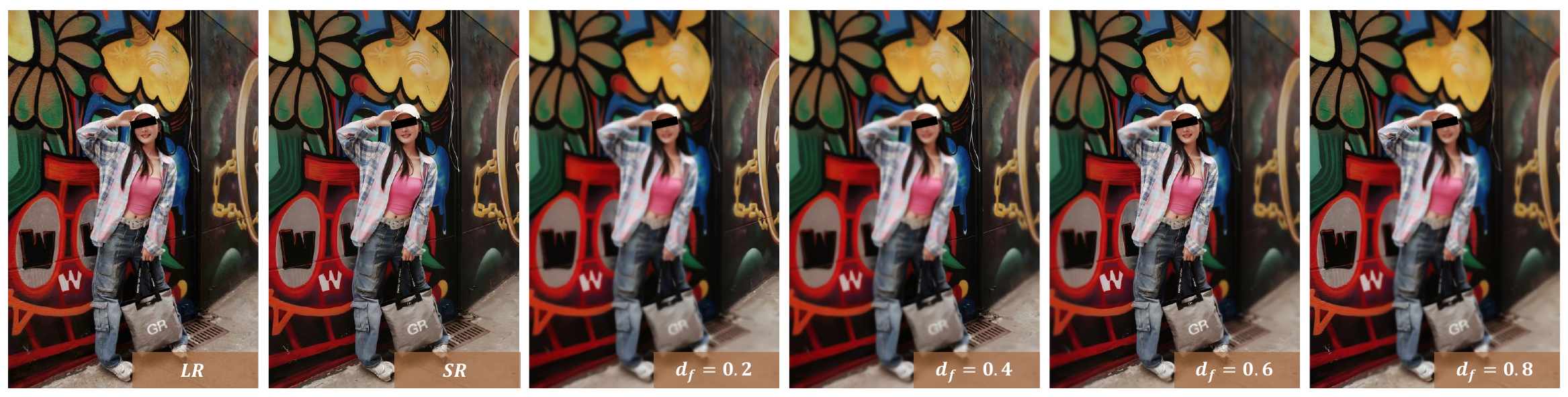}
  \vspace{-0.7cm}
  \caption{Given the disparity map and LR input, our method is able to achieve dynamic adjustment of the focus distance.}
  \vspace{-0.2cm}
  \label{fig:focus}
\end{figure*}

\begin{figure*}[!ht]
  \centering
  \includegraphics[width=\linewidth]{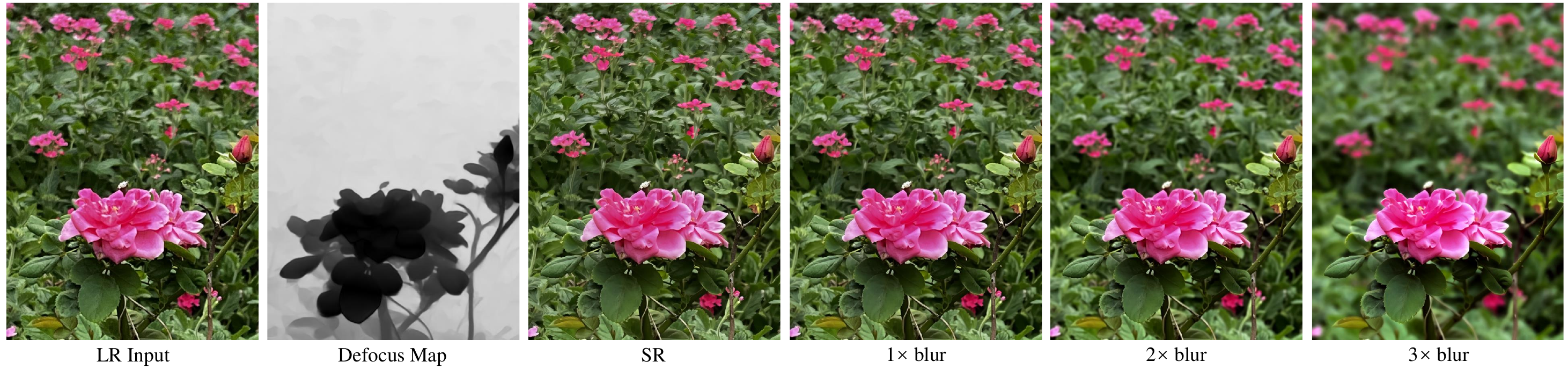}
  \vspace{-0.7cm}
  \caption{Given the defocus map and LR input, our method is able to gradually increase the aperture parameter from 1x blur to 3x blur.}
  \vspace{-0.4cm}
  \label{fig:aper}
\end{figure*}

\section{Quantitative comparison on Real-ISR}

Although our method is not specifically designed for super-resolution tasks, setting the blur intensity $K$ to 0 allows us to obtain all-in-focus HR images. Furthermore, despite not incorporating text conditions, MagicBokeh still shows performance in the single task of Real-ISR, as illustrated in Tab. \ref{table:5}, highlighting its ability to restore both the realism and aesthetic quality of images.

\section{More Results}
\subsection{Adjusting Focus Distance}
We provide examples of changing focus distance in Fig. \ref{fig:focus}. Whether focusing on the foreground or background, our method achieves natural super-resolution and bokeh effects.

\subsection{Adjusting Aperture}
We present the results of increased blurriness in Fig. \ref{fig:aper}. MagicBokeh successfully achieves progressive blurriness while maintaining subject sharpness. The cases are high-zoom real mobile device captures, and MagicBokeh generates realistic bokeh effects.

\begin{figure*}[t]
  \centering
  \includegraphics[width=0.9\linewidth]{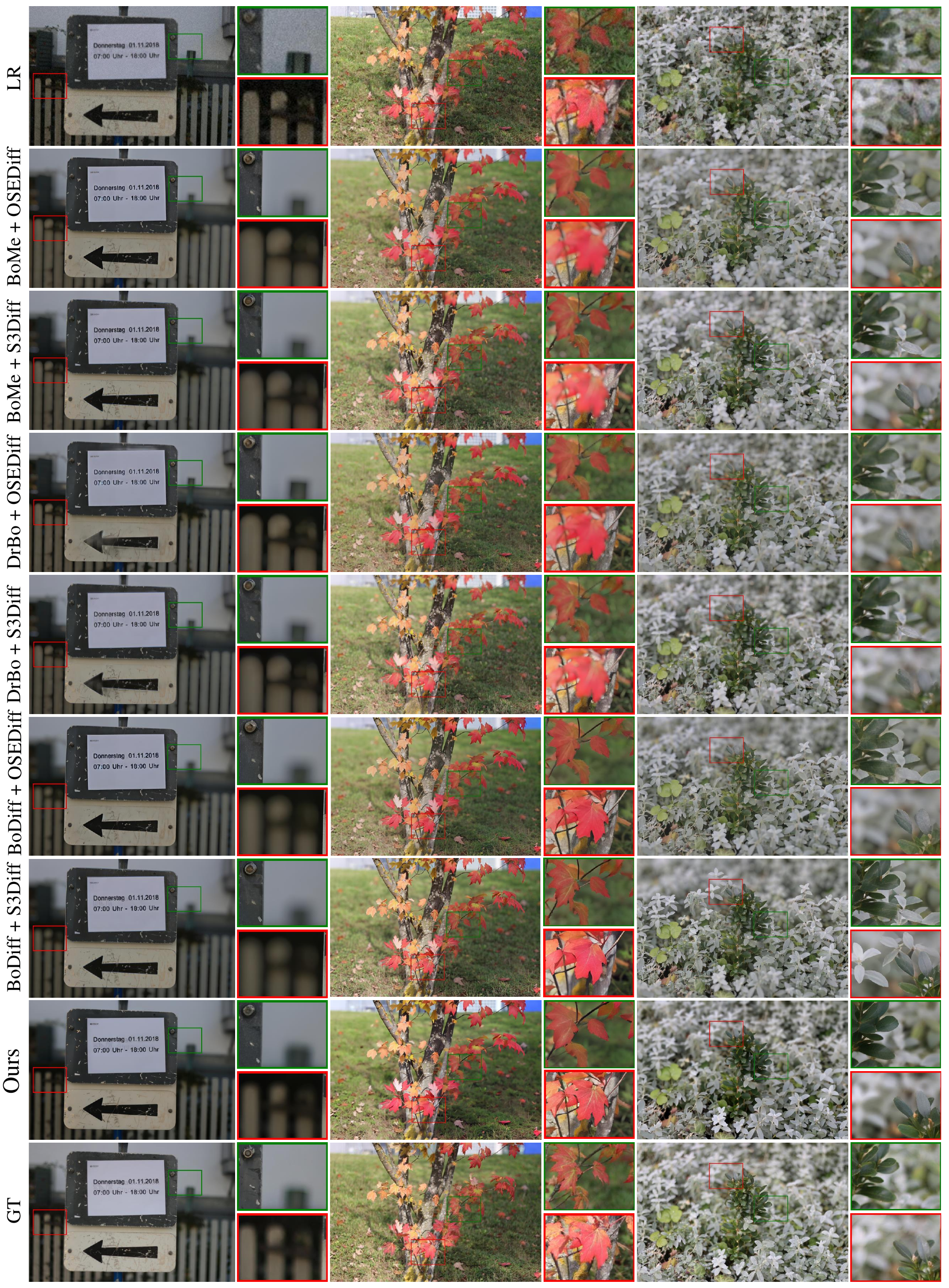}
  \caption{Qualitative comparison on EBB400-LQ (\textit{Zoom-in for best view}).}
  \label{fig:comp1}
\end{figure*}

\begin{figure*}[t]
  \centering
  \includegraphics[width=0.9\linewidth]{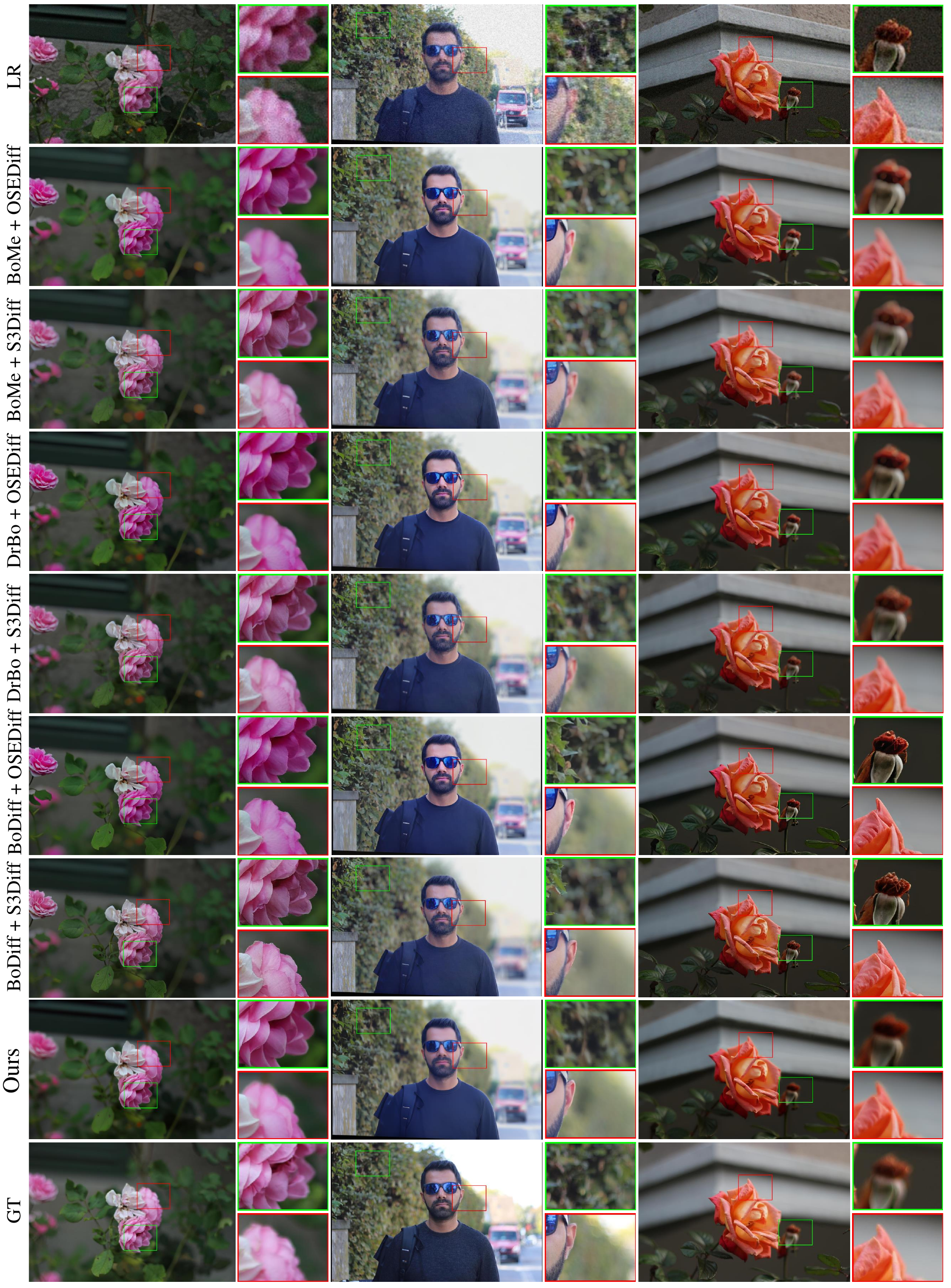}
  \caption{Qualitative comparison on EBB400-LQ (\textit{Zoom-in for best view}).}
  \label{fig:comp2}
\end{figure*}

\subsection{More Comparisons}
\vfill
Here, we provide more comparisons between MagicBokeh and other two-stage pipeline to further validate the effectiveness of MagicBokeh. First, we demonstrate more comparisons in Fig. \ref{fig:comp1}. In the first example, MagicBokeh produces bokeh effects that are closer to the Ground Truth compared to other methods, especially in the red-boxed area. Compared to methods including BokehMe and Dr.Bokeh in the green-boxed area, our method and BokehDiff generate sharper edges. In the second example, in terms of super-resolution, our method produces more distinct leaf details compared to OSEDiff and S3Diff. In terms of bokeh, our method generates the best edge effects compared to BokehDiff, BokehMe, and Dr.Bokeh. In the third example, our method can still produce bokeh effects that are consistent with the real situation, even in the presence of noise. We continue the results demonstration in Fig. \ref{fig:comp2}. Our method gradually increases the blur with increasing defocus while keeping the focused foreground unchanged, resulting in a more realistic effect.

{
    \small
    \bibliographystyle{ieeenat_fullname}
    \bibliography{main}
}

\end{document}